\documentclass[a4paper, 10pt, conference]{IEEEtran}
\IEEEoverridecommandlockouts
\usepackage{cite}
\usepackage{amsmath,amssymb,amsfonts}
\usepackage{algorithmic}
\usepackage{algorithm}
\usepackage{comment}
\makeatletter
\let\MYcaption\@makecaption
\makeatother
\usepackage{svg}
\usepackage{subcaption}
\makeatletter
\let\@makecaption\MYcaption
\makeatother

\usepackage{graphicx}
\usepackage{textcomp}
\usepackage{xcolor}
\def\BibTeX{{\rm B\kern-.05em{\sc i\kern-.025em b}\kern-.08em
    T\kern-.1667em\lower.7ex\hbox{E}\kern-.125emX}}

\DeclareMathOperator*{\maxmize}{\text{maxmize}}
\DeclareMathOperator*{\minimize}{\text{minimize}}
\newcommand{\vect}[1]{\boldsymbol{#1}}

\graphicspath{{./figs/}}
\begin{document}

\title{Adversarial joint attacks on legged robots\\
\thanks{This work was supported by JSPS KAKENHI Grant Numbers JP19K12039 and JP22H03658.}
}

\author{\IEEEauthorblockN{1\textsuperscript{st} Takuto Otomo}
\IEEEauthorblockA{\textit{Graduate School of Science and Engineering} \\
\textit{Chiba University}\\
Chiba, Japan \\
Email: takutootomo@chiba-u.jp}
\and
\IEEEauthorblockN{2\textsuperscript{nd} Hiroshi Kera}
\IEEEauthorblockA{\textit{Graduate School of Engineering} \\
\textit{Chiba University}\\
Chiba, Japan \\
Email: kera@chiba-u.jp}
\and
\IEEEauthorblockN{3\textsuperscript{rd} Kazuhiko Kawamoto}
\IEEEauthorblockA{\textit{Graduate School of Engineering} \\
\textit{Chiba University}\\
Chiba, Japan \\
Email: kawa@faculty.chiba-u.jp}
}

\maketitle

\begin{abstract}
We address adversarial attacks on 
the actuators at the joints of legged robots trained by deep reinforcement learning.
The vulnerability to the joint attacks can significantly impact
the safety and robustness of legged robots.
In this study, 
we demonstrate that the adversarial perturbations
to the torque control signals of the actuators
can significantly reduce the rewards and cause walking instability in robots.
To find the adversarial torque perturbations,
we develop black-box adversarial attacks, where, the adversary 
cannot access the neural networks trained by deep reinforcement learning.
The black box attack can be applied to legged robots regardless of the architecture and
algorithms of deep reinforcement learning.
We employ three search methods for the black-box adversarial attacks: random search, differential evolution, and numerical gradient descent methods.
In experiments with the quadruped robot Ant-v2 and 
the bipedal robot Humanoid-v2, in OpenAI Gym environments,
we find that differential evolution can efficiently find
the strongest torque perturbations among the three methods.
In addition, we realize that the quadruped robot Ant-v2
is vulnerable to the adversarial perturbations, whereas
the bipedal robot Humanoid-v2 is robust to the
perturbations.
Consequently, the joint attacks can be used for proactive diagnosis of robot 
walking instability.
\end{abstract}


\section{Introduction}
% Since legged robots require complex, multi-degree-of-freedom control, acquisition of control laws by deep reinforcement learning is considered suitable(e.g.\cite{9560769}).
With the recent progress in deep reinforcement learning for robot control, safety and robustness have become the primary concern.
This concern is especially true for legged robots
because they are prone to falling \cite{9560769}. 
Many disturbances can affect the safety and robustness of legged robots.
Among disturbances, one should pay attention to adversarial attacks
\cite{szegedy2014intriguing,goodfellow2015explaining} because these attacks 
can significantly impact the safety and robustness with small perturbations.
For legged robots,
finding vulnerability to adversarial attacks leads to proactive detection of potential risks of falling.

This study deals with the adversarial attacks on the control of legged robots in deep reinforcement learning.
In particular, we demonstrate that perturbations to the torque control signals of actuators at joints, which are essential components for walking, cause instability in walking motion.
We use an adversarial attack~\cite{szegedy2014intriguing,goodfellow2015explaining}
to search such adversarial perturbations.
Adversarial attacks in supervised learning seek input perturbations that reduce the loss for deep reinforcement learning, whereas, in this study, we seek torque perturbations that reduce the reward.
The torque perturbations are possibly caused by 
the actuator's modeling error, noise, or physical failure
\cite{9322775},
rather than intentional attacks by the malicious 
adversary.
Thus, finding vulnerability to the torque perturbations
can be used to proactively diagnose robot-walking instability.

Many adversarial attacks in deep reinforcement learning perturb the state observations to cause the agents to malfunction
\cite{xiao2019characterizing,9536399}.
The attacks on the state observations allow
\textit{white-box attacks} where the adversary
can access the neural networks, such as
policy networks and Q-networks, that take the state
observations as the input~\cite{huang2017,3238064,NEURIPS2020_f0eb6568}.
%The white-box attacks can efficiently generate
%strong attacks.
Conversely, we consider a \textit{black-box attack} on the joint actuators of legged robots by perturbing the torque signals.
This attack is defined in the action space.
The adversarial attacks on the action space 
are usually black-box ones because
the environment functions are neither neural networks nor
differentiable functions, though
white-box attacks are possible using proxy reward functions~\cite{lee2020spatiotemporally}.
For legged robots, the environment functions are
physical simulators or real-world trials.
In addition,
the black box attack can be applied to 
robots regardless of the architectures and algorithms 
of deep reinforcement learning.
\begin{figure}[t]
    \centering
    \includegraphics[width=\linewidth]{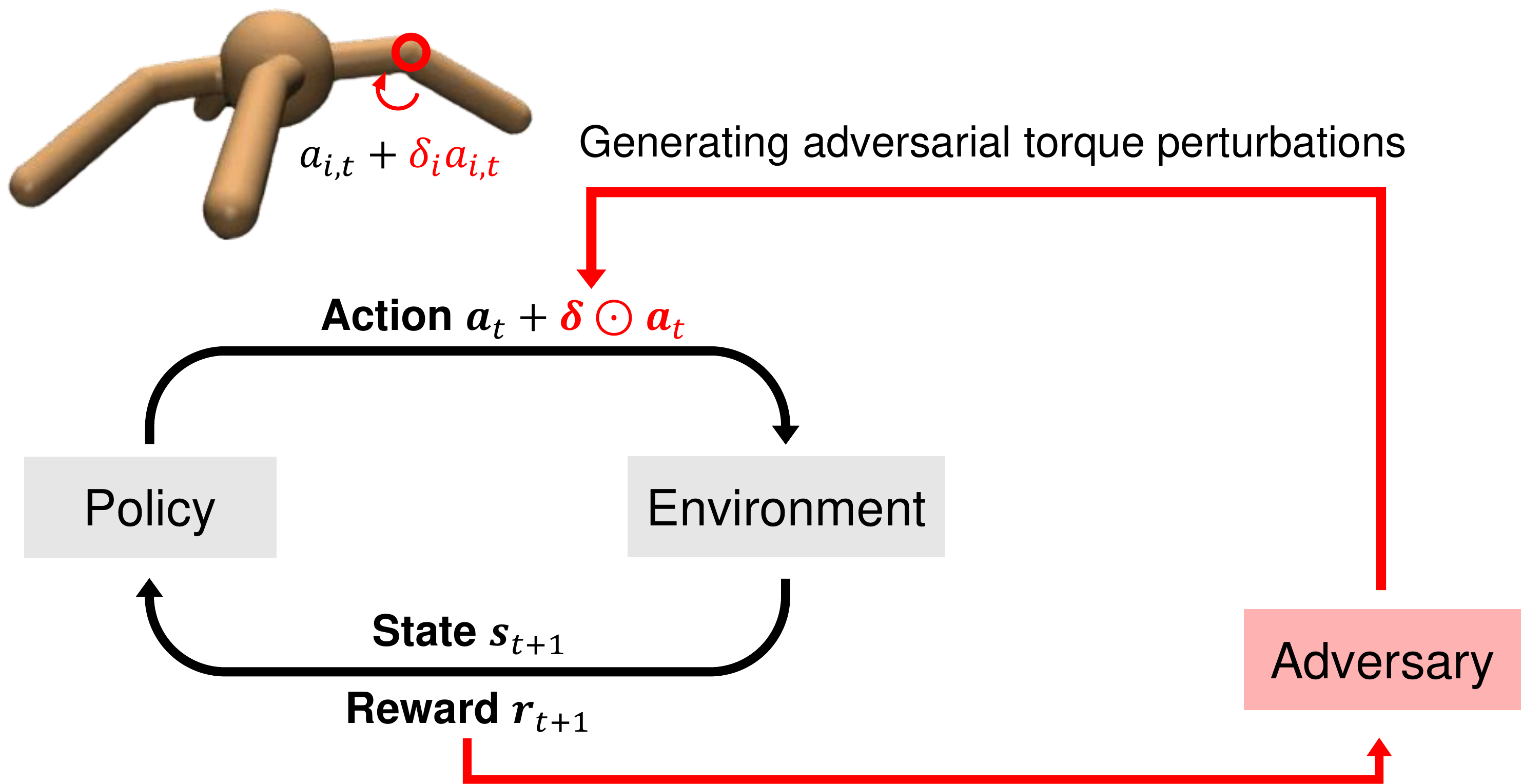}
    \caption{Overall view of adversarial joint attack}
    \label{overall}
\end{figure}

We introduce three search methods for black-box
adversarial joint attacks to legged robots:
random search, differential evolution~\cite{storn1997differential}, and numerical gradient 
descent methods.
These three methods generate torque perturbations that reduce
the rewards by repeating walking simulations of legged robots.
The torque perturbations are assumed to be fixed over time.
%Therefore, we need to repeat the walking simulation to search for adversarial perturbations. Considering that different search methods can produce different convergence, we tried the random search method, differential evolution method, and gradient methods.
%Our attack method is a black-box attack because the structure and parameters of the deep model are not assumed.
We conduct experiments in the quadruped robot Ant-v2 and 
the bipedal robot Humanoid-v2 environments
from OpenAI Gym\cite{brockman2016openai}. 
These environments run on the
physical engine MuJoCo \cite{todorov2012mujoco}.
We evaluate the effectiveness of the three search methods 
in terms of the average cumulative rewards.
The experimental results reveal that the differential evolution method
can efficiently detect the strongest torque perturbations
among the three methods.
In addition, we discover that the quadruped robot Ant-v2
is vulnerable to the adversarial perturbations, whereas
the bipedal robot Humanoid-v2 is robust to the
perturbations.

The key contributions of this study can be summarized as follows:
\begin{itemize}
  \item We propose a black-box adversarial joint attack on legged robots. 
  \item We demonstrate that the differential evolution method can efficiently find the strongest torque perturbations among
  the three methods.
  \item We discover  the torque perturbations that interfere with the walking task for the first time.
\end{itemize}
\begin{figure}[t]
    \centering
    \includegraphics[width=\linewidth]{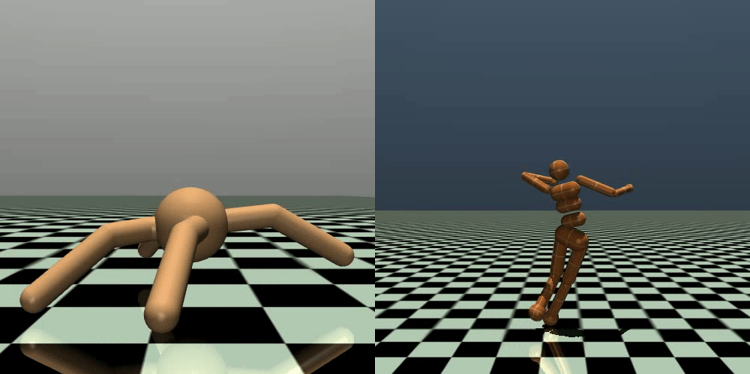}
    \caption{Legged robot: Left Ant-v2, Right Humanoid-v2}
    \label{walker}
\end{figure}
\section{RELATED WORK}
Adversarial attacks \cite{goodfellow2015explaining} on image classification in supervised learning  apply small perturbations $\vect{\delta}_x$ to the input image $\vect{x}$ to misclassify the deep model.
Such perturbations $\vect{\delta}_x$ are sought to maximize the loss function.
\begin{align}
    \maxmize_{\vect{\delta}_x}
    J(\vect{x}+\vect{\delta}_x),\ \text{subject to}\
    ||\vect{\delta}_x||_p \le \epsilon,
    \label{equ: adv_attack}
\end{align}
where, $J(\cdot)$ is the loss function and $||\cdot||_p$ is the $p$-norm.
For deep models, it is challenging to find the optimal solution of Eq. (\ref{equ: adv_attack}).
As an approximate solution method, the gradient ascent method using the gradient of the loss function $\nabla_{\vect{x}} J(\vect{x})$ with respect to the input image $\vect{x}$ is used \cite{goodfellow2015explaining,madry2018towards}.
To calculate the gradient $\nabla_{\vect{x}} J(\vect{x})$, the structure and parameters of the deep model are 
assumed to be known, and such an attack is called a white box attack.

Adversarial attacks on deep reinforcement learning are often formulated to minimize the expected reward.
Because there are multiple attack targets (such as states, actions, environments, and rewards) various attacks are possible in
deep reinforcement learning \cite{9536399}.
A well-known adversarial attack on TV gameplay \cite{huang2017} 
is defined on the state space, which attacks adds an adversarial perturbation $\vect{\delta_x}$ to the input TV screen $\vect{x}$.
This attack is formulated the same as the optimization problem in Eq.~(\ref{equ: adv_attack}) because the input is an image and the output action is discrete.
There are several adversarial attacks against robot control with continuous states and actions such as those proposed in this study.
One approach is to introduce an enemy agent that attacks the legged robot, and the two play against each other to obtain robust policies \cite{pinto2017robust,pmlr-v97-tessler19a,Gleave2020Adversarial}.
These methods attack the legged robot from the outside.
Unlike those in this study, they do not assume any failures or modeling errors in the legged robot itself.
For attacks in action space, similar to those in this study, Lee~et~al.~\cite{lee2020spatiotemporally} conducted a white-box attack on the proxy reward function to find adversarial perturbations that vary in space and time.
Conversely, we assume physical failures and modeling errors and seek black-box time-fixed adversarial perturbations.
Yang~et.~al~\cite{9588329} used the greedy method to search for adversarial failures, assuming failures that cause joints to stop moving altogether while acquiring robust measures by training on such failures.
In this study, we do not consider binary expressions of the presence or absence of a fault but small torque perturbations.

\section{Simulation-based Adversarial Joint Attack}
We introduce three adversarial attack methods for finding joint torque perturbations that can interfere with the walking movements of legged robots.
First, we describe adversarial attacks on the action space and define joint torque perturbations. Next, we describe a simulation-based search to find the adversarial perturbation.

\subsection{Adversarial attacks on joints}\label{AA}
In this study, we attack the legged robots by perturbing the action vector $\vect{a}_t\in\mathcal{R}^{N_a}$,
which is stochastically generated by a policy network $\pi(\vect{a}_t|\vect{s}_t)$  with a continuous state vector $\vect{s}_t\in\mathcal{R}^{N_s}$.
The action vector $\vect{a}_t$ is the torque control signal of the joint actuators at time $t$,
where, $\mathcal{R}$ is the real number set, and $N_a$ is the total number of the joint actuators.
We denote the adversarial perturbation
to the action vector $\vect{a}_t$ by $\vect{\delta}$.
We assume that the perturbation $\vect{\delta}$ is fixed over time and it is multiplicative, i.e.,
the attacked action vector $\vect{a}'_t$ is given by
\begin{equation}
  \vect{a}'_t=\vect{a}_t+\vect{\delta}\odot\vect{a}_t,\
  \text{where}\ ||\vect{\delta}||_\infty\le\epsilon,
  \label{equ: perturbation}
\end{equation}
where, $\odot$ is the Hadamard product, which is the product of each component of the vectors.
For the $i$-th torque control signal $a_{i,t}$
of $\vect{a}_t$, the signal is perturbed by
\begin{equation}
a'_{i,t} = a_{i,t} + \delta_{i} a_{i,t}
=(1+\delta_{i})a_{i,t}.
\end{equation}
The time-fixed perturbation is possibly caused by either the actuator's modeling error or physical failure,
which both remain nearly fixed over time.

We seek the adversarial torque perturbations $\vect{\delta}$ 
by minimizing the expected cumulative reward as
\begin{align}
    \minimize_{\vect{\delta}}\ E[C(\vect{\delta})],\ \text{subject to}\
    ||\vect{\delta}||_\infty \le \epsilon,
    \label{equ: adv_attack_action}
\end{align}
where, $C(\vect{\delta})$ is the cumulative reward 
attacked by the adversarial perturbation $\vect{\delta}$
and is defined as
\begin{align}
    C(\vect{\delta})=\sum_{t=0}^{T-1}R(\vect{s_{t}},\vect{a}_t+\vect{\delta}\odot\vect{a}_t),
    \label{equ: cum_reward}
\end{align}
where, $R(\vect{s_{t}},\vect{a}_t)$ is the reward at time $t$ given state $\vect{s_{t}}$ and action $\vect{a}_t$.
The expected value in Eq.~(\ref{equ: adv_attack_action}) is taken with respect to 
a trajectory of states and actions $\{\vect{s}_0,\vect{a}_0,\vect{s}_1,\vect{a}_1,\ldots,\vect{s}_{T-1},\vect{a}_{T-1}\}$.

\subsection{Simulation-based Black-Box Search}
We consider the black-box adversarial attack,
where, we cannot access the policy network $\pi(\vect{a}_t|\vect{s}_t)$ and we can only
access the rewards $R(\vect{s_{t}},\vect{a}_t)$
through the waling simulator.
In this case, it is difficult to compute the gradient
$\nabla_{\vect{a}} C(\vect{\delta})$
of the cumulative reward in Eq.~(\ref{equ: adv_attack_action})
with respect to the action vector $\vect{a}$.
Therefore, it is necessary to search for the adversarial perturbations by repeating walking simulations,
that is, we empirically 
estimate the expected cumulative reward
$E[C(\vect{\delta})]$
in Eq.~(\ref{equ: adv_attack_action})
by taking the average of $M$ cumulative rewards as
\begin{equation}
E[C(\vect{\delta})]\approx \bar{C}(\vect{\delta})=
\frac{1}{M}\sum_{m=1}^MC^{(m)}(\vect{\delta}),
\label{equ: empirical cumulative reward}
\end{equation}
where, $C^{(m)}(\vect{\delta})$ is a cumulative reward of the $m$-th
walking simulation.
For each simulation run,
the initial state $\vect{s}_0$ is 
reset before the simulation.
In the experiments, we set $M=100$.
We employ three simulation-based search
methods that minimize the empirical estimation 
in Eq.~(\ref{equ: empirical cumulative reward}): random search, differential evolution, 
and numerical gradient descent methods.

\subsubsection{Random Search}
We represent the random search method as
\textbf{Algorithm} \ref{alg1}.
The random search method randomly
generates a set of torque perturbations 
$\vect{\delta}^{(n)},n=1,2,\ldots$.
Each element of $\vect{\delta}^{(n)}$ is 
generated according to the uniform distribution
$U([-\epsilon, \epsilon])$.
For each perturbation $\vect{\delta}^{(n)}$, we run $M$ walking simulations 
and compute the average cumulative reward
using Eq.~(\ref{equ: empirical cumulative reward}).
Then, we choose the best perturbation 
$\vect{\delta}_{\mathrm{best}}$ that minimizes
the average cumulative reward
as the adversarial perturbation.

\begin{algorithm}[t]
    \caption{Random Search for an Adversarial Attack}
    \label{alg1}
    \begin{algorithmic}[1]
    \STATE $C_{\mathrm{min}} \leftarrow \infty$
    \FOR {each perturbation $n=1,2,\ldots$ }
    \STATE Generate torque perturbation $\vect{\delta}^{(n)}\sim U\left([-\epsilon,\epsilon]^{N_a}\right)$
	\STATE Compute average cumulative reward $\bar{C}(\vect{\delta}^{(n)})$\! in Eq.\!~(\ref{equ: empirical cumulative reward})
	\IF {$\bar{C}(\vect{\delta}^{(n)}) < C_{\mathrm{min}}$}
    \STATE $C_{\mathrm{min}} \leftarrow \bar{C}(\vect{\delta}^{(n)})$
    \STATE Update best torque perturbation $\vect{\delta}_\mathrm{best} \leftarrow \vect{\delta}^{(n)}$
	\ENDIF
    \ENDFOR
    \end{algorithmic}
\end{algorithm}

\subsubsection{Differential Evolution}
We represent the differential evolution method
as \textbf{Algorithm} \ref{alg2}.
For differential evolution,
we have a population of torque perturbations
$\{\vect{\delta}_g^{(1)},\ldots,\vect{\delta}_g^{(NP)}\}$, where, $NP$ is the population size and the subscript
$g$ stands for the $g$-th generation.
Each element of the initial individual $\vect{\delta}_0^{(i)}$ 
is generated according to the uniform distribution
$U\left([-\epsilon,\epsilon]\right)$.
We use the average cumulative reward 
in Eq.~(\ref{equ: empirical cumulative reward})
to evaluate the fitness of each individual
$\vect{\delta}_g^{(i)}$.
The best individual 
$\vect{\delta}_{\mathrm{best}}$ is one
that yields the lowest average cumulative
reward, not the highest.

We use the mutation and crossover processes for
differential evolution as follows:
For mutation,
we choose the best/1 strategy and 
obtain the $i$-th mutant individual $\vect{v}_{g+1}^{(i)}$
as
\begin{equation}
%\label{best1my}
\vect{v}_{g+1}^{(i)} = \vect{\delta}_{\mathrm{best}} + F(\vect{\delta}_g^{(r_1)} - \vect{\delta}_g^{(r_2)}),
\label{equ: mutation}
\end{equation}
where,
$F \in(0.5,1]$ is a random coefficient,
$r_1$ and $r_2$ are mutually exclusive
indices randomly chosen from 
$\{1,2,\ldots,NP\}$.
For crossover,
we use the binomial crossover and obtain 
the $j$-th element of the $i$-th trial
individual $\vect{u}_{g+1}^{(i)}$ as 
\begin{equation}
{u}_{j,g+1}^{(i)} = \left\{
\begin{array}{ll}
{v}_{j,g+1}^{(i)} & \text{if}\quad r \leq CR\ \text{or}\ j=j_{r}\\
{\delta}_{j,g}^{(i)} & \text{otherwise}
\end{array}
\right.
\label{equ: crossover}
\end{equation}
where, 
$r$ is a uniform random number in $[0,1]$,
$CR=0.7$ is a crossover constant,
and $j_r$ is an index randomly chosen from
$[1,2,\ldots,N_a]$.
We use the clip function
to keep the perturbation
within the range $[-\epsilon,\epsilon]$
element-wise.
The clip function is defined by
$\mathrm{clip}_{[-\epsilon,\epsilon]}(\vect{x})=\max(\min(\vect{x},\epsilon),-\epsilon)$,
where, $\max$ and $\min$ are taken element-wise.
Finally we choose the best individual
$\vect{\delta}_{\mathrm{best}}$
from all generation populations.

\begin{algorithm}[t]
    \caption{Differential Evolution for an Adversarial Attack}
    \label{alg2}
    \begin{algorithmic}[1]
    \REQUIRE Population size $NP$
    \STATE Generate initial population of torque perturbations $\{\vect{\delta}_0^{(1)},\ldots,\vect{\delta}_0^{(NP)}\}$ with $\vect{\delta}_0^{(i)}\sim U\left([-\epsilon,\epsilon]^{N_a}\right)$ 
    \STATE $C_{\mathrm{min}} \leftarrow \infty$
    \FOR {each generation $g = 1,2,\ldots$}
	\FOR {each individual $i = 1,2,\ldots,NP$}
	    \STATE Generate trial individual $\vect{u}_{g}^{(i)}$ by mutation and crossover in Eqs.~(\ref{equ: mutation}) and (\ref{equ: crossover})
	    \STATE $\vect{u}_{g}^{(i)} \leftarrow \mathrm{clip}_{[-\epsilon,\epsilon]}(\vect{u}_{g}^{(i)})$
    	\STATE Compute average cumulative rewards in Eq.~(\ref{equ: empirical cumulative reward}) for\\
    	trial $\bar{C}_(\vect{u}_g^{(i)})$
        and target $\bar{C}(\vect{\delta}_{g-1}^{(i)})$
        \IF {$\bar{C}(\vect{u}_g^{(i)}) \leq \bar{C}(\vect{\delta}_{g-1}^{(i)})$}
        \STATE Accept trial individual $\vect{\delta}_{g}^{(i)} \leftarrow \vect{u}_{g}^{(i)}$
        \IF {$\bar{C}(\vect{u}_g^{(i)})<C_{\mathrm{min}}$}
        \STATE $C_{\mathrm{min}}\leftarrow \bar{C}(\vect{u}_g^{(i)})$
        \STATE Update best individual $\vect{\delta}_{\mathrm{best}}\leftarrow \vect{\delta}_g^{(i)}$
        \ENDIF
	\ELSE
        \STATE Retain target individual $\vect{\delta}_{g}^{(i)} \leftarrow \vect{\delta}_{g-1}^{(i)}$ 
	\ENDIF
	\ENDFOR
    \ENDFOR
    \end{algorithmic}
\end{algorithm}

\subsubsection{Numerical Gradient Descent}
As mentioned previously, 
it is difficult to compute the gradient vector
$\vect{g}=\nabla_{\vect{a}} C(\vect{\delta})$
of the cumulative reward for gradient descent.
Therefore, we numerically approximate the
gradient vector $\vect{g}$
using a finite difference method,
that is,
the $i$-th element of $\vect{g}$ is approximated by
\begin{equation}
    g_i=\frac{\partial C(\vect{\delta})}{\partial a_i}
    \approx \frac{C(\vect{\delta}+\vect{h}_i)-C(\vect{\delta})}{h},
    \label{equ: finite difference}
\end{equation}
where, $h$ is a finite difference and $\vect{h}_i$ is defined as a 
vector whose $i$-th element is $h$ and the rest are zeros, i.e.,
$\vect{h}_i = (0,\ldots, h, \ldots, 0)$.
$g_i$ is the gradient with respect
to the torque signal of $i$-th actuator.

We represent the numerical gradient descent
method as \textbf{Algorithm} \ref{alg3}.
This algorithm terminates at a predetermined search time.
In the algorithm, 
we use the clip function
to keep the perturbation $\vect{\delta}$
within the range $[-\epsilon,\epsilon]$
element-wise.

\begin{algorithm}[t]
    \caption{Gradient Descent for an Adversarial Attack}
    \label{alg3}
    \begin{algorithmic}[1]
    \STATE Initialize torque perturbation $\vect{\delta}\leftarrow (0,0,\ldots,0)$
    \STATE $C_{\mathrm{min}} \leftarrow \infty$
    \WHILE{termination criterion is not met}\STATE Compute average cumulative reward $\bar{C}(\vect{\delta})$\! in Eq.\!~(\ref{equ: empirical cumulative reward})
    \IF{$\bar{C}(\vect{\delta})< C_{\mathrm{min}}$}
     \STATE $C_{\mathrm{min}}\leftarrow \bar{C}(\vect{\delta})$
     \STATE Update best torque perturbation $\vect{\delta}_\mathrm{best}\leftarrow\vect{\delta}$
    \ENDIF
    \FOR{each actuator $i = 1,2,\ldots,N_a$}
	\STATE Compute gradient $g_i$ in Eq.~(\ref{equ: finite difference})
    \ENDFOR
    \STATE Normalize gradient $\vect{g}\leftarrow \vect{g}/||\vect{g}||_\infty$
    \STATE $\vect{\delta} \leftarrow \mathrm{clip}_{[-\epsilon,\epsilon]}(\vect{\delta}-\alpha\vect{g})$
    \ENDWHILE
    \end{algorithmic}
\end{algorithm}

\section{EXPERIMENTS with legged robots}
This experiment aims to search for small perturbations to leg joints of legged robots, which interfere with the walking movements, and consequently find  the most vulnerable joint.
To this end, we compare the three search
methods in terms of search capability and
attack strength.

\subsection{Experiment setup}
We conduct experiments in the quadruped
robot Ant-v2 and the bipedal robot Humanoid-v2 environ-
ments from OpenAI Gym, as illustrated in Fig.~\ref{walker}.
Ant-v2 and Humanoid-v2 have eight and seventeen joints, respectively.
The simulation environment is built using the MuJoCo physics engine \cite{todorov2012mujoco}.
The task is forward walking, and the legged robot is trained to walk sufficiently using proximal policy optimization (PPO)\cite{schulman2017proximal} in advance.

The reward function for Ant-v2 is defined by \begin{equation}
  \label{reward}
	R(\vect{s},\vect{a})=v_{\text{fwd}}-0.5\|\vect{a}\|^{2}-0.5\cdot10^{-3}\left\|\vect{f}\right\|^{2}+1,
\end{equation}
and for Humanoid-v2 by
\begin{equation}
         \label{hum_reward}
	R(\vect{s},\vect{a})=3v_{\text{fwd}}-0.1\|\vect{a}\|^{2}-0.5\cdot10^{-6}\left\|\vect{f}\right\|^{2}+5,
\end{equation}
where, $v_{\text{fwd}}$ is the forward velocity,
$\vect{a}$ is the torque signal vector,
and $\vect{f}$ is the 
impact forces~\cite{rewardfunction}.

\subsection{Comparison of search capability}
\begin{figure}[t]
  \begin{center}
  \includegraphics[width=\linewidth]{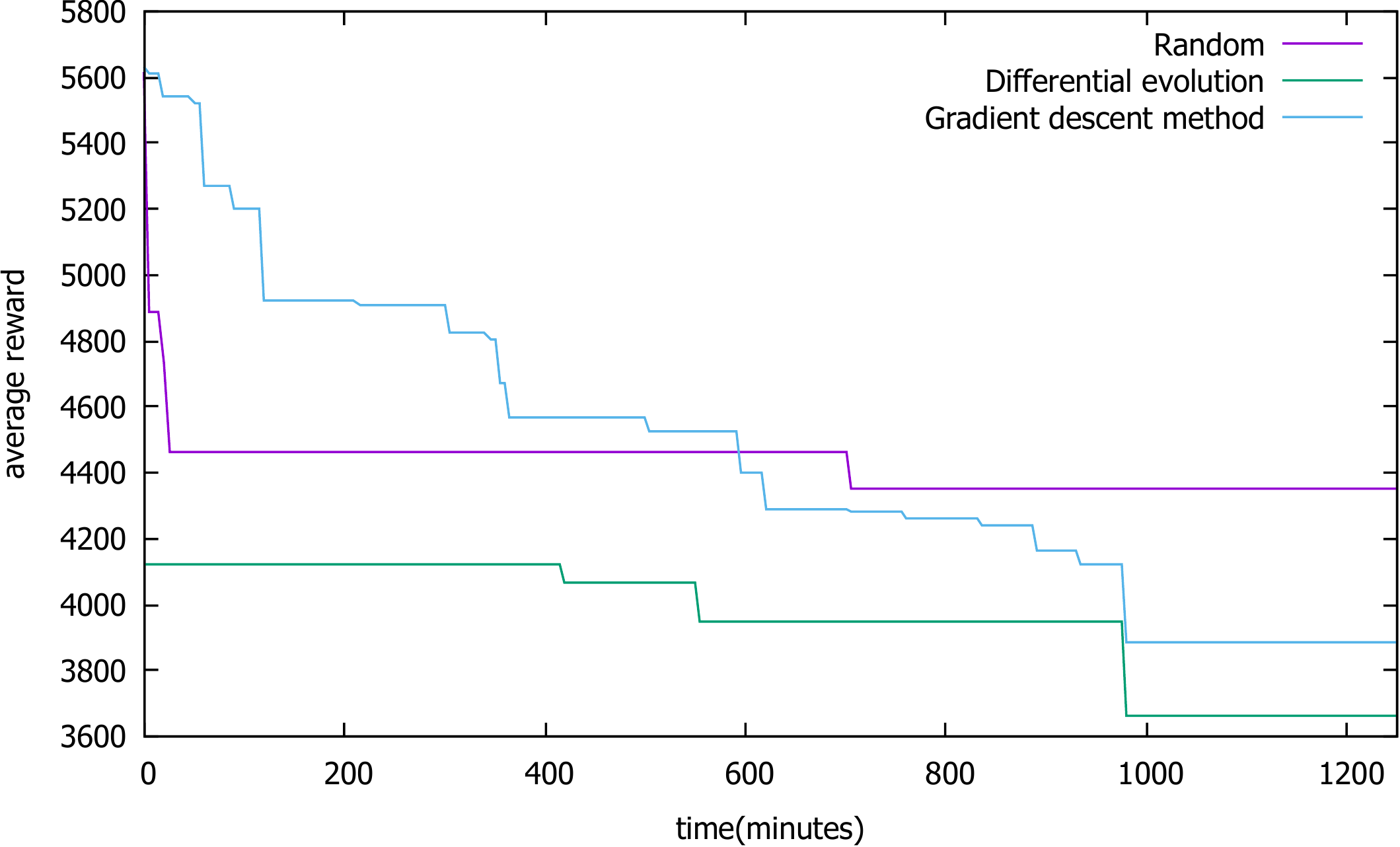}
  \end{center}
  \caption{Average cumulative rewards by
  three search methods (Ant-v2)}
  \label{ant_comp}
\end{figure}

\begin{figure}[t]
  \begin{center}
  \includegraphics[width=\linewidth]{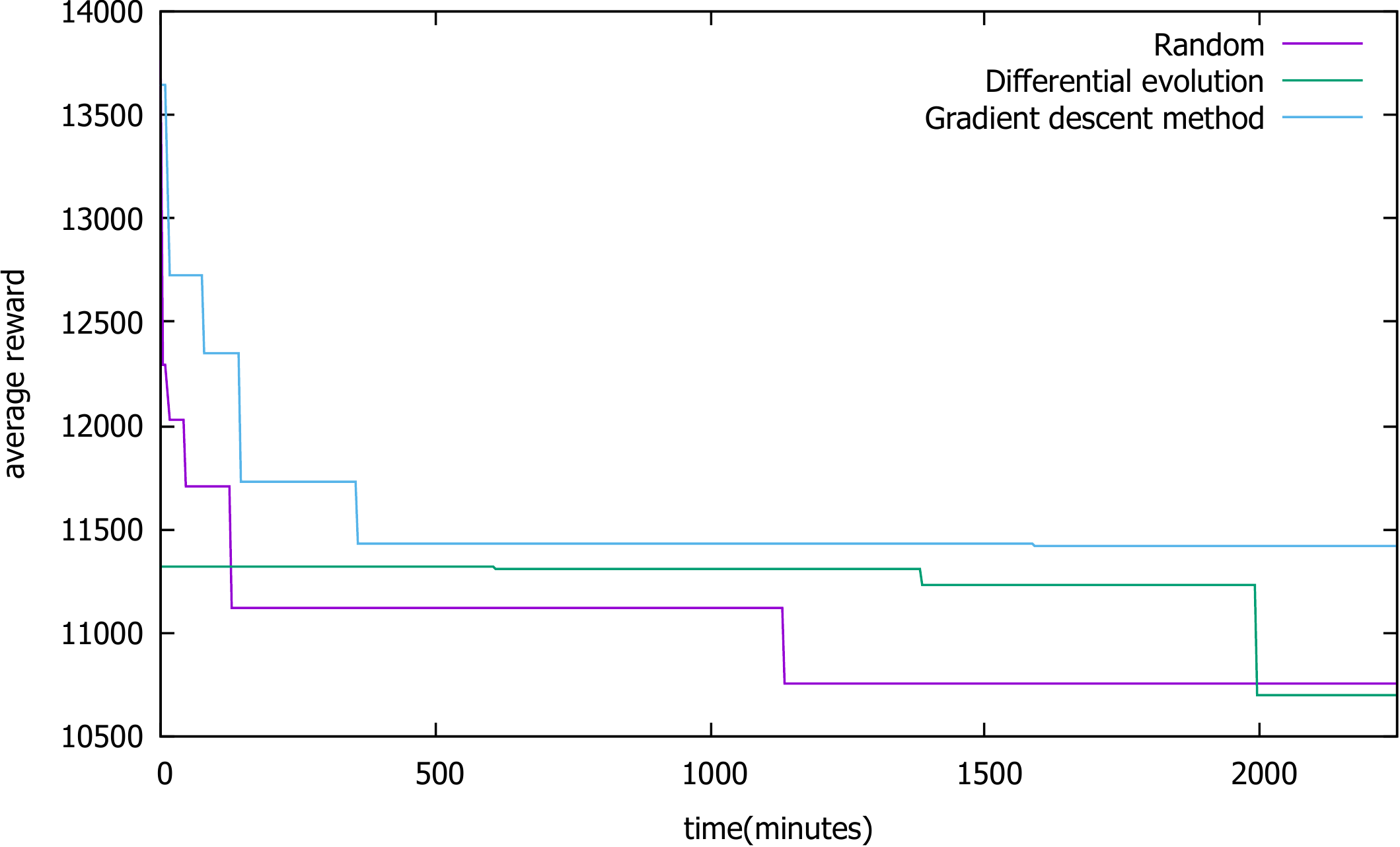}
  \end{center}
  \caption{Average cumulative rewards by
  three search methods (Humanoid-v2)}
  \label{hum_comp}
\end{figure}

Fig. \ref{ant_comp} and \ref{hum_comp}
illustrate the average cumulative rewards
$\bar{C}(\vect{\delta})$
of the best perturbation  $\vect{\delta}_{\mathrm{best}}$
for random search (purple line), differential evolution (green line), and gradient descent (cyan line).
For a fair comparison, the search time is set to 1250 min for Ant-v2 and 2250 min for Humanoid-v2.
We plot the average cumulative rewards
every five min in Figs.~\ref{ant_comp} and \ref{hum_comp}.

From Fig.~\ref{ant_comp},
the differential evolution method (green line) can find the strongest attack parameters than the other two methods for Ant-v2.
In addition, the differential evolution method stably converges and, it has the highest convergence speed.
Conversely,
Fig.~\ref{hum_comp} demonstrate that the differential
method are comparable to the random search for Humanoid-v2.
As mentioned below, Humanoid-v2 is robust to the adversarial attack.
Therefore, the search methods cannot find effective perturbations.

\begin{comment}
\begin{figure}[t]
  \begin{center}
  \includegraphics[width=\linewidth]{ant_pgd_comp.png}
  \end{center}
  \caption{Comparison of the convergence of gradient methods by minute change range (Ant-v2)}
  \label{ant_pgd_comp}
\end{figure}
\end{comment}

\subsection{Adversarial attack by differential evolution}
As reported above, the differential evolution
method can find the best adversarial
perturbation among the three methods.
This subsection evaluates
the adversarial attacks by 
the differential evolution method.
In experiments, we set the number of individuals
to $NP=120$ for Ant-v2 and $NP=255$ for Humanoid-v2.

\begin{figure}[t]
  \begin{center}
  \includegraphics[width=0.85\hsize]{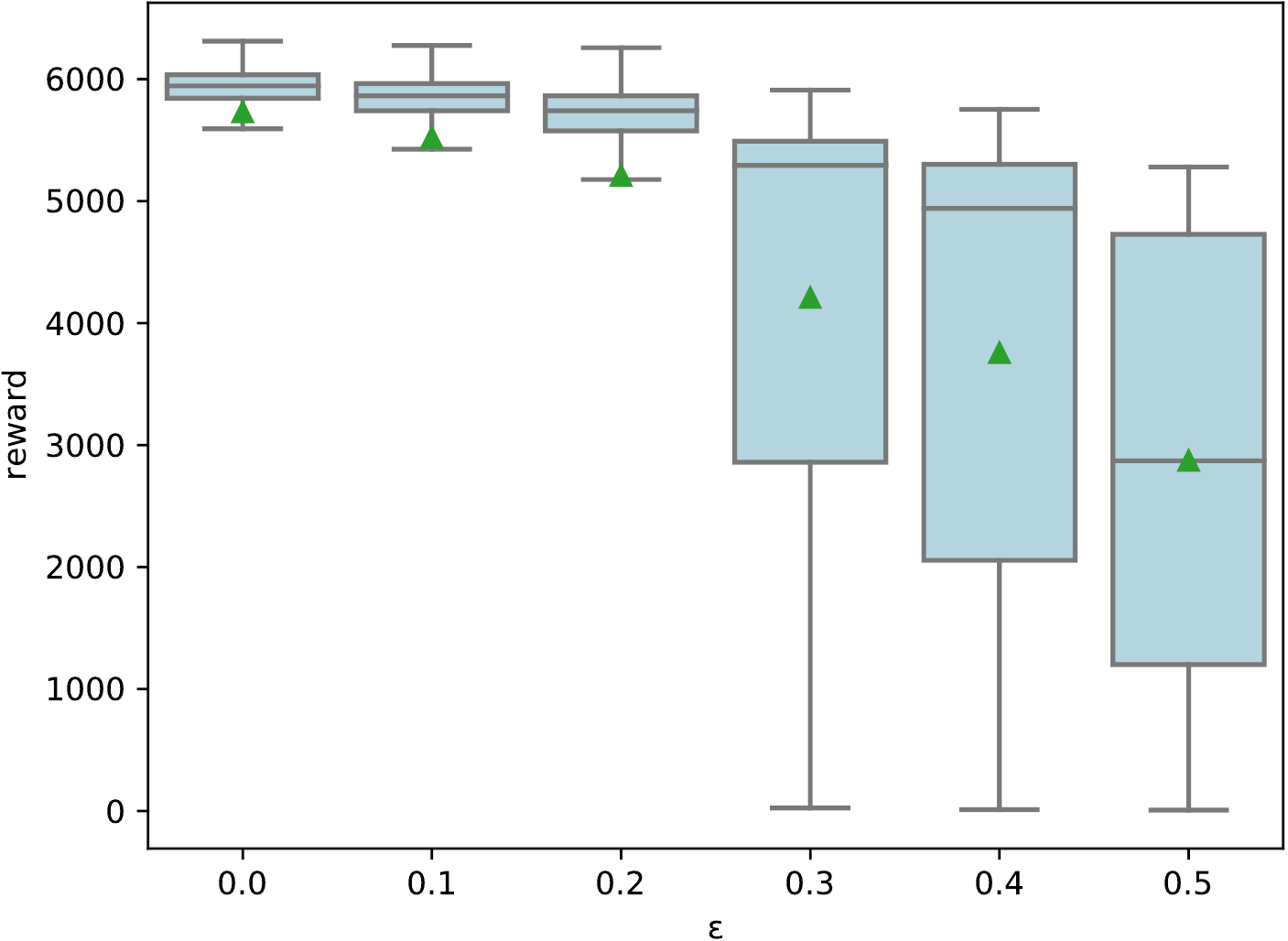}
  \end{center}
  \caption{Cumulative reward distribution
  by differential evolution with
  varying attack strength 
  $\epsilon$ (Ant-v2)}
  \label{hakoant}
\end{figure}
\begin{figure}[t]
  \begin{center}
  \includegraphics[width=0.85\hsize]{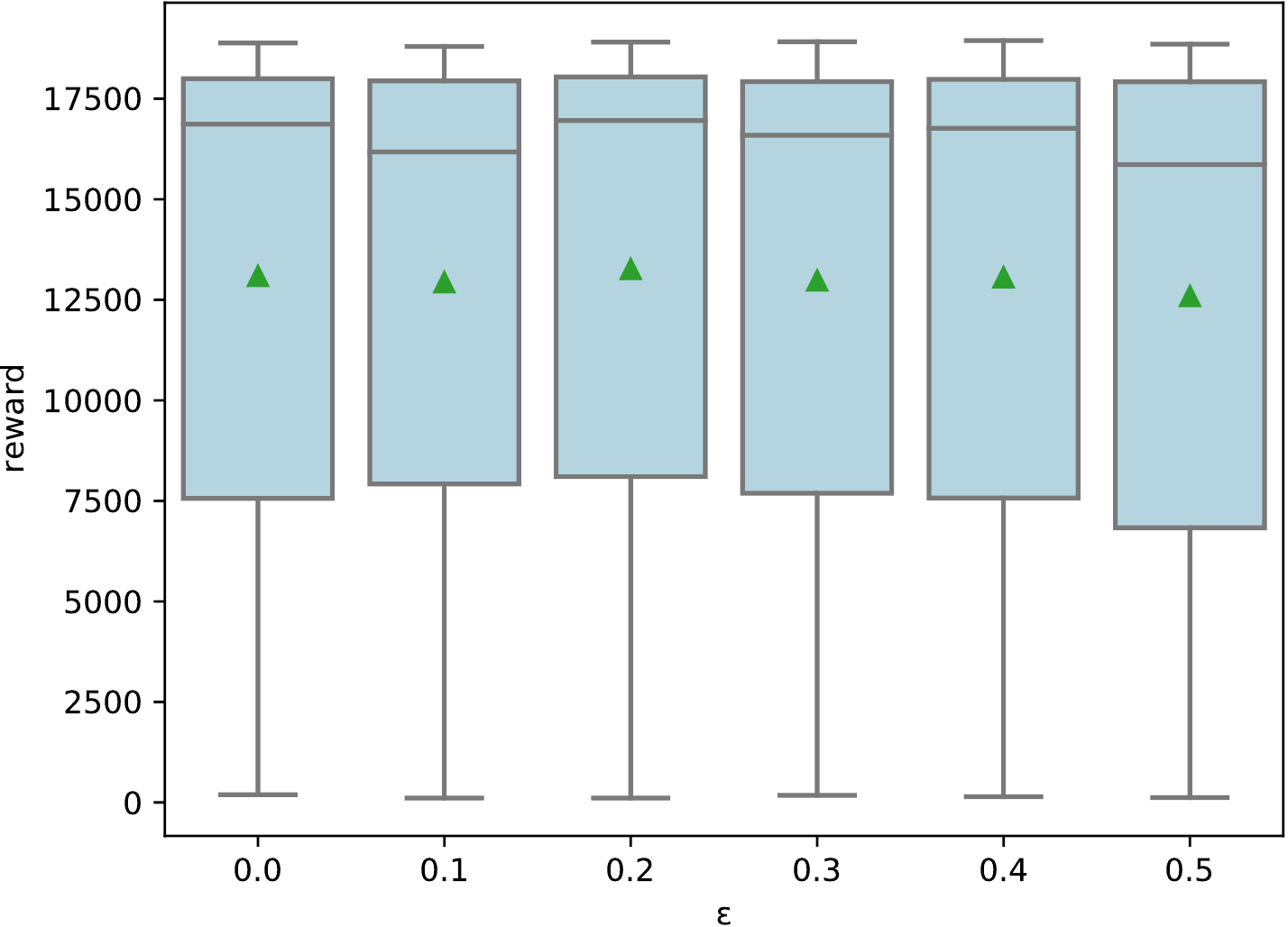}
  \end{center}
  \caption{Cumulative reward distribution
  by differential evolution with
  varying attack strength 
  $\epsilon$ (Humanoid-v2)}
  \label{hakohum}
\end{figure}

We test how the cumulative rewards are affected by the varying attack strength $\epsilon=0.0,0.1,0.2,0.3,0.4,0.5$.
$\epsilon=0.0$ means no attack.
Figs. \ref{hakoant} and \ref{hakohum} illustrate
the box plots of the cumulative rewards by
1000 simulations.
The green triangles in the plots are the average 
cumulative rewards.
From Fig.~\ref{hakoant},
we observe that the attacks to Ant-v2 with $\epsilon=0.3$ and greater significantly reduce the cumulative rewards.
These results indicate that 
Ant-v2 is vulnerable to the adversarial attacks.
Conversely, 
from Fig.~\ref{hakohum},
we observe that the attacks on Humanoid-v2 do not
change the cumulative rewards regardless of the 
attack strengths.
Therefore we conclude that Humanoid-v2 is robust to the adversarial attacks.
Unlike Ant-v2, which can stably walk with four legs, Humanoid-v2 has experienced many falls during learning and has learned how not to fall,
thereby Humanoid-v2 became robust.
Another possible explanation is that, 
the effect of the joint torques is not easily transmitted to walking
because the feet in contact with the ground are fewer than in the case of Ant-v2. 

\begin{figure*}[!t]
% \begin{tabular}{c}
  \begin{minipage}[t]{85mm}
  \centering
  \includegraphics[width=\linewidth]{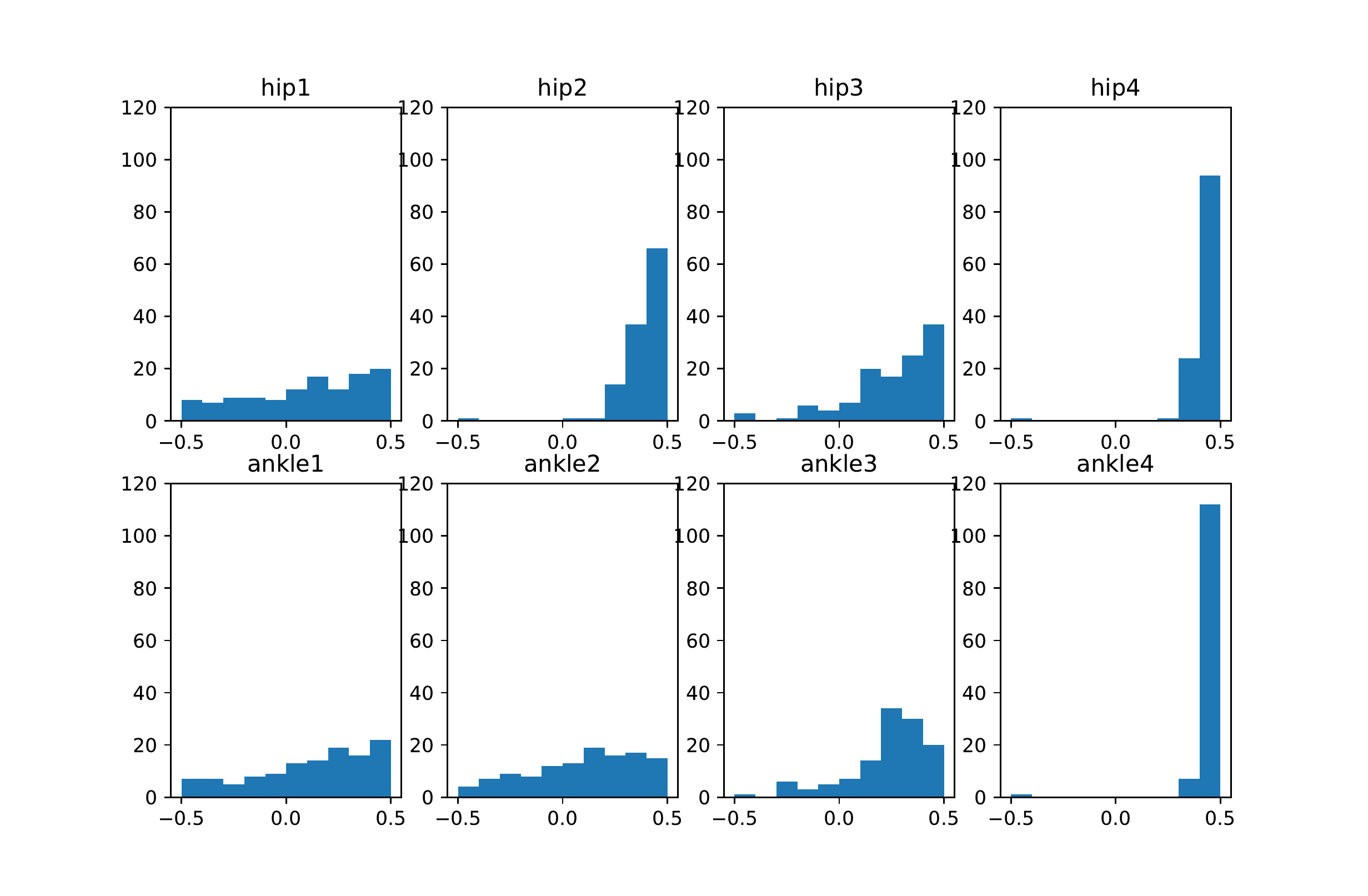}
  \subcaption{Attack strength distributions per actuator}
  \label{ant_part_hist}
  \end{minipage}
  \begin{minipage}[t]{85mm}
  \centering
  \includegraphics[width=\linewidth]{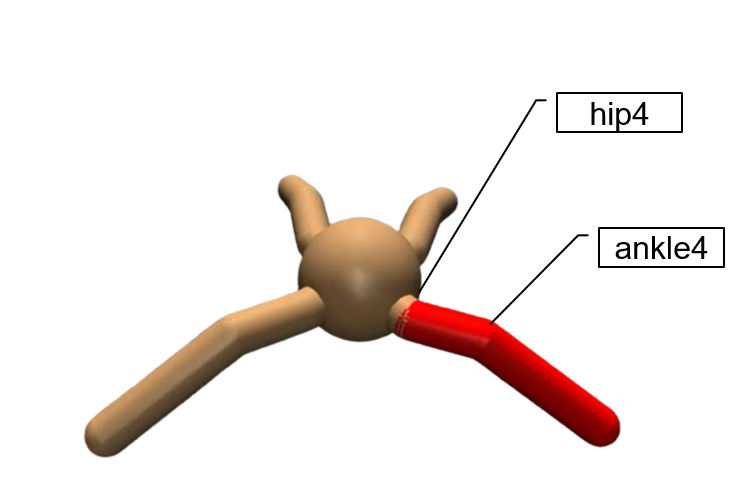}
  \subcaption{Most attacked actuators}
  \label{ant_part}
  \end{minipage}
%   \end{tabular}
  \caption{Adversarial attacks on Ant-v2 for 
  last generation individuals of differential evolution}
  \label{ant_part_hikaku}
\end{figure*}

\begin{table}[t]
    \caption{Adversarial best torque perturbation for Ant-v2 ($\epsilon = 0.5$)}
    \label{ratio ant}
    \centering
    \begin{tabular}{|l||r|} \hline
    	body part		& perturbation $\delta$\\ \hline \hline
    	hip1 			 & \color{red}-0.01 		\\ %\hline
    	ankle1 			  &  \color{blue}+0.09 		\\ \hline
    	hip2 			 &  \color{blue}+0.44 		\\ %\hline
    	ankle2 			 &  \color{blue}+0.35  		\\ \hline
        hip3 			 &  \color{blue}+0.33		\\ %\hline
    	ankle3 			 &  \color{blue}+0.45  		\\ \hline
    	hip4 			 &  \color{blue}+0.49 		\\ %\hline
    	ankle4 			 & \color{blue}+0.47 	 	\\ \hline
    \end{tabular}
\end{table}

\begin{figure*}[!t]
\centering
\subfloat[Locomotion with normal (no attack)]{\includegraphics[clip, width=\linewidth]{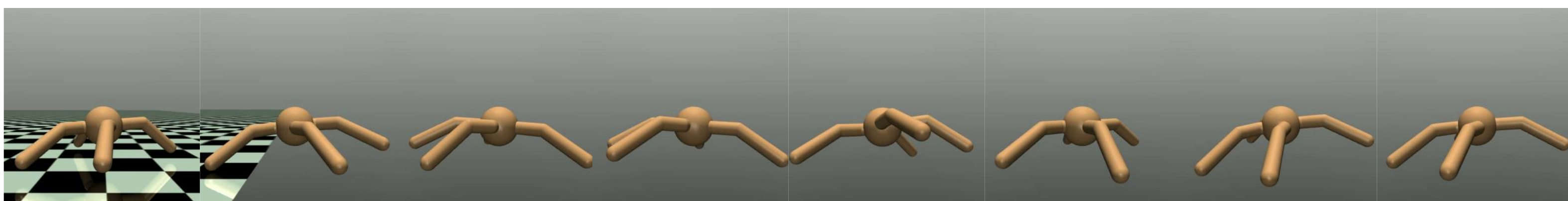}
\label{fig:label-A}}
\\
\subfloat[Locomotion with adversarial joint attack]{\includegraphics[clip, width=\linewidth]{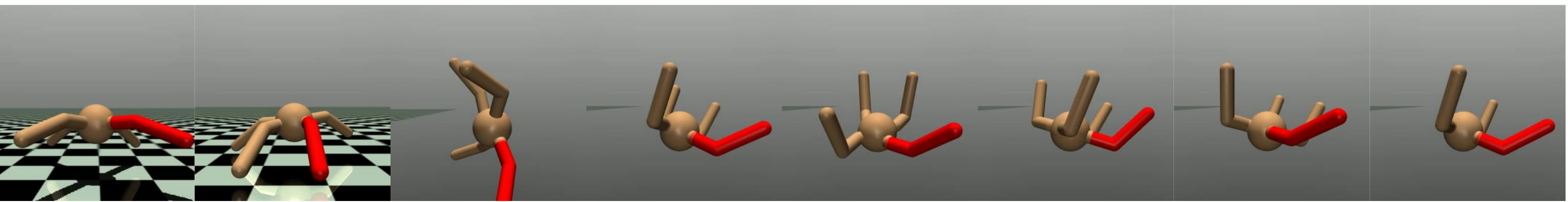}
\label{fig:label-B}}
\\
\caption{Illustration of quadruped robot locomotion for Ant-v2}
\label{locomotion}
\end{figure*}

We further investigate the attacks on Ant-v2 using
the strongest attack strength $\epsilon=0.5$.
Table \ref{ratio ant} shows the best perturbation $\vect{\delta}_\mathrm{best}$ by the differential evolution method with
$\epsilon = 0.5$. From the table, we observe that the attack strongly 
perturbs the three legs (hip2, ankle2, hip3, ankle3, hip4, and ankle4),
and does not do the other leg (hip1 and ankle1).
Thus, the attack unbalances the walking motion.
Fig. \ref{ant_part_hist} illustrates the histograms of
the attack strengths of 120 individuals at the last generation
for each actuator of Ant-v2.
From Fig.~\ref{ant_part_hist}, we observe that 
the individuals mostly attack the actuators 
at hip4 and ankle4 
of Ant-v2, as illustrated in Fig.~\ref{ant_part}.
Observing the walking simulation animation, as shown in the Fig.~\ref{locomotion} (bottom), 
we find that these actuators are located at the primary leg for walking.
The attacks on these actuators increase the magnitude of the
torque signals and add an excessive vertical force.
Consequently, Ant-v2 tends to fall as shown in the Fig.~\ref{locomotion}.
% \begin{figure}[t]
%   \begin{center}
%   \includegraphics[width=1.0\hsize]{attack.png}
%   \end{center}
%   \caption{Attack in $\epsilon$}
%   \label{hako}
% \end{figure}

\section{CONCLUSION}
This study investigates the vulnerability
to the adversarial joint attacks on the quadruped
and bipedal robots trained by deep reinforcement learning.
Deep reinforcement learning has been becoming popular
for complex multi-degree-of-freedom control such as legged-robots control.
Thus, it will increasingly become important to diagnose 
the vulnerability, safety, and robustness
for real-world applications.
This study demonstrates that the legged robot
can be vulnerable to joint attacks and can be forced to fall.

In the experiments, we observed that differential evolution can efficiently find the strongest torque perturbations.
Interestingly, the quadruped robot Ant-v2 is vulnerable and the bipedal robot Humanoid-v2 is robust.
For future work, we will develop defense methods
against adversarial joint attacks to improve the 
safety and robustness of legged robots.

\bibliographystyle{unsrt}
\bibliography{refs}
% \section{Appendices}
% \begin{table}[H]
%     \caption{Adversarial torque perturbation to Ant-v2 ($\epsilon = 0.5$)}
%     \label{ratio humanoid}
%     \centering
%     \begin{tabular}{|l||r|} \hline
%     	body part		& perturbation $\delta$\\ \hline \hline
%     	hip1 			 & \color{red}-0.01 		\\ %\hline
%     	ankle1 			  &  \color{blue}+0.09 		\\ \hline
%     	hip2 			 &  \color{blue}+0.44 		\\ %\hline
%     	ankle2 			 &  \color{blue}+0.35  		\\ \hline
%         hip3 			 &  \color{blue}+0.33		\\ %\hline
%     	ankle3 			 &  \color{blue}+0.45  		\\ \hline
%     	hip4 			 &  \color{blue}+0.49 		\\ %\hline
%     	ankle4 			 & \color{blue}+0.47 	 	\\ \hline
%     \end{tabular}
% \end{table}
\end{document}